%% file: main.tex
\documentclass[10pt,twocolumn,letterpaper]{article}

\usepackage[pagenumbers]{iccv} 

\input{preamble}

\usepackage[ruled,vlined]{algorithm2e}  
\usepackage{amsmath}  
\usepackage{tcolorbox}

\definecolor{iccvblue}{rgb}{0.21,0.49,0.74}
\usepackage[pagebackref,breaklinks,colorlinks,allcolors=iccvblue]{hyperref}

\title{Human + AI for Accelerating Ad Localization Evaluation}

\author{
Harshit Rajgarhia \quad Shivali Dalmia \quad Mengyang Zhao \quad Abhishek Mukherji \quad Kiran Bharadwaj Ganesh\\
Centific Global Solutions Inc.\\
{\tt\small harshit.rajgarhia@centific.com}
}

\usepackage{graphicx}
\usepackage{float}

\begin{document}
\maketitle
\input{sec/0_abstract}    
\input{sec/1_introY}

\input{sec/3_approachY}

\input{sec/3.5_resultY}

\input{sec/4_conclusionY}

{\small
\bibliographystyle{IEEEtran}
\bibliography{main}
}


\end{document}

%% file: preamble.tex
\usepackage{float}


%% file: sec/0_abstract.tex
\begin{abstract}
Adapting advertisements for multilingual audiences requires more than simple text translation; it demands preservation of visual consistency, spatial alignment, and stylistic integrity across diverse languages and formats. We introduce a structured framework that combines automated components with human oversight to address the complexities of advertisement localization. To the best of our knowledge, this is the first work to integrate scene text detection, inpainting, machine translation (MT), and text reimposition specifically for accelerating ad localization evaluation workflows. Qualitative results across six locales demonstrate that our approach produces semantically accurate and visually coherent localized advertisements, suitable for deployment in real-world workflows.
\vspace{-3mm}
\end{abstract} 

%% file: sec/1_introY.tex
\section{Introduction}
\vspace{-1mm}
Given the strong focus on being relatable to the target audience to maximize product sales, advertisement (Ad) localization plays a critical role in ensuring that a product is linguistically and culturally tailored to align with the preferences and norms of the target locale \cite{kanso2002advertising, taylor1996french, tai1997advertising}. Companies often navigate their corporate identities by aligning advertising strategies with the cultural and geographical landscapes of their markets. Machine translation (MT) has significantly advanced in handling text and speech modalities, as evidenced by \cite{kocmi2024preliminarywmt24rankinggeneral,ahmad2024findingsiwslt2024evaluation}. However, there has been growing interest within the machine learning community in exploring translation in the visual modality \cite{caglayan2019probing}.

In industrial settings, the application and the evaluation of translated content are typically carried out by human translation experts, especially in the visual modality. Due to the lack of automated evaluation mechanisms, traditional image analysis workflows rely heavily on manual processing, which are time-consuming and expensive \cite{khanuja-etal-2025-towards, zhao2021faster}. Previous studies have mostly focused on individual components—e.g., OCR, MT evaluation, and image manipulation~\cite{8991281,baek2019characterregionawarenesstext,banerjee-lavie-2005-meteor}—but no unified system exists for streamlining Ad localization at scale.

\begin{figure*}
    \centering
    \includegraphics[width=1.0\linewidth]{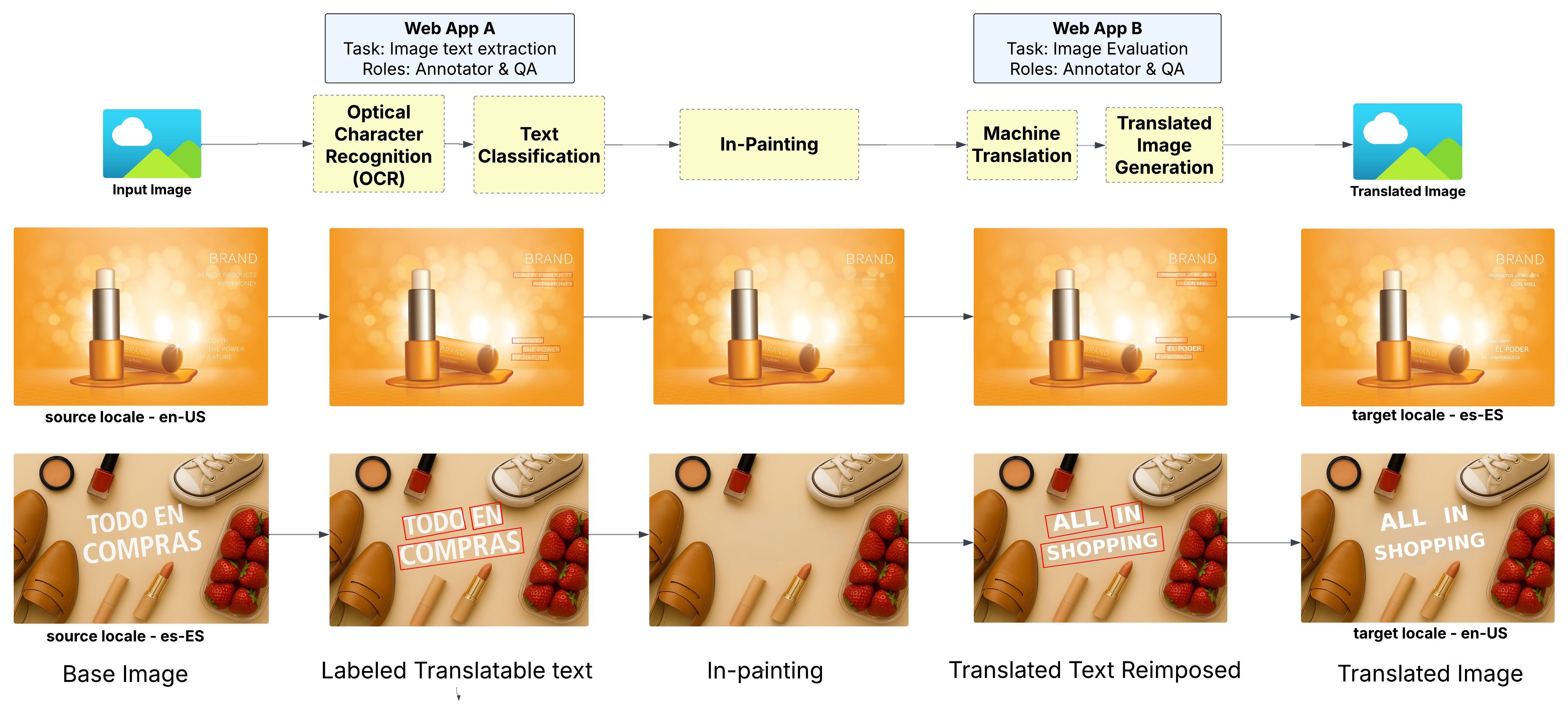}
    \caption{Ad Localization Evaluation Process}
    \label{fig:Architecture}
    \vspace{-6mm}
\end{figure*}

Our goal is to fill this gap by offering a practical, end-to-end framework, including text recognition, inpainting, and reimposition, which is suitable for industrial deployment. In this paper, we develop an Ad localization system based on deep learning and large language models to address the key bottlenecks of traditional workflows. The proposed system breaks down MT-based Ad localization into a stepwise pipeline (as shown in Fig.~\ref{fig:Architecture}), and explores how each step can leverage AI solutions to improve overall quality and efficiency. Our major contributions include:
\vspace{1mm}
\begin{itemize}
    \item Developed a deep learning-based scene text detection engine with a human-in-the-loop mechanism, significantly improving accuracy and reducing average annotation time from 40 minutes to 15 minutes.
    \item Built an automated pipeline using Stable Diffusion to regenerate masked text regions and selectively recombine them for seamless and scalable background restoration, eliminating the need for manual retouching.
    \item Implemented a geometry and typography aware reimposition module using OpenCV-based rendering to accurately place translated text, minimizing manual effort while preserving visual fidelity.
    \item Demonstrated that the AI-driven automation reduces overall handling time from hours—or even days—to just minutes.
\end{itemize}

%% file: sec/3_approachY.tex
\section{Approach}

This section outlines the scope, high-level architecture (See Fig. \ref{fig:Architecture}), and key components of the pipeline such as text extraction and categorization, in-painting, machine translation, and text reimposition. This work outlines the design and implementation of an automated pipeline for the localization and translation of the advertisements image for 6 locales (en-ES, it-IT, fr-CA, sv-SV, nl-NL, en-US) as listed in Table 1. 
\vspace{-1mm}

\begin{table}[h]
\centering
\resizebox{\columnwidth}{!}{%
\begin{tabular}{|l|c|c|c|}
\hline
\textbf{Source Locale} & \textbf{Target Locale} & \textbf{Training volumes}\\
\hline
es-US (Spanish - United States) & en-US (English - United States) & 50+  \\\hline
en-US (English - United States) & es-US (Spanish - United States) & 50+ \\\hline
en-US (English - United States) & fr-CA (French - Canada) & 50+\\\hline
it-IT (Italian - Italy) & en-US (English - United States) & 50+ \\\hline
fr-CA (French - Canada) & en-US (English - United States) & 50+ \\\hline
sv-SE (Swedish - Sweden) & en-US (English - United States) & 50+\\\hline
nl-NL (Dutch - Netherlands) & en-US (English - United States) & 50+ \\\hline
\hline
\end{tabular}%
}
\caption{Number of advertisement Images per \textit{Locale} per \textit{month}}
\label{tab:localevolumes}
\vspace{-5mm}
\end{table}
\subsection{Architecture }
Our ad localization system follows a modular, step-wise workflow consisting of text annotation, background inpainting, machine translation, and localized text reimposition. Input images are uploaded to cloud storage and accessed via a custom web interface (Web App A), where annotators classify each detected text region into one of three categories: Brand, Translatable, or Certification.

Text detection is performed by an integrated OCR model triggered on task load. Annotators refine bounding boxes and labels, after which the image proceeds through automated inpainting and translation modules. Translated text is re-rendered with attention to typographic and spatial attributes extracted from the original layout.

Final outputs are reviewed via Web App B, which displays the source and localized images side-by-side for evaluation. Annotators assess visual coherence, translation accuracy, and rendering quality. Quality assurance reviewers make final approval decisions to close the loop.

\subsection{Optical Character Recognition} 
\label{sec:ocr-arch}
We adopt EasyOCR~\cite{easyocr} for multilingual scene text detection and recognition. It combines the CRAFT model~\cite{baek2019characterregionawarenesstext} for region-level text detection and a Convolutional Recurrent Neural Network (CRNN) for character recognition. The CRNN includes CNN-based spatial feature extraction followed by LSTM layers and a Connectionist Temporal Classification (CTC) decoder for robust sequence prediction.

Prior to recognition, input images undergo preprocessing steps such as despeckling, binarization, contrast enhancement (CLAHE), and resizing, which improve text visibility in cluttered ad backgrounds. CRAFT generates character region and affinity score maps to segment coherent text groups, while the CRNN maps these into structured Unicode sequences.

This combination enables accurate extraction of diverse fonts and orientations, supporting over 80 languages. Its flexibility and robustness make it particularly well-suited for visually complex advertisements.

\begin{figure*}[t]
  \centering
  \includegraphics[width=0.45\linewidth]{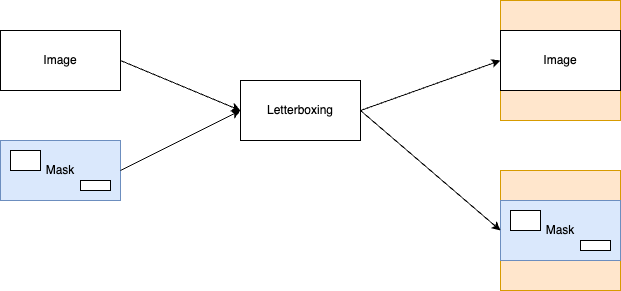}
  \hfill
  \includegraphics[width=0.45\linewidth]{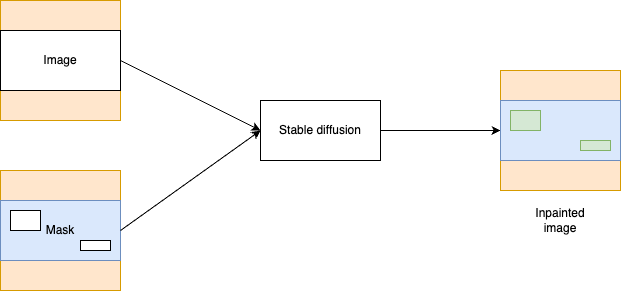}
  \caption{
    Inpainting workflow: (1.)  
    Letter boxing. (2.) Inpainting.
  }
  \label{fig:inpainting_workflow}
  \vspace{-4mm}
\end{figure*}

\begin{figure*}[t]
  \centering
  \includegraphics[width=0.45\linewidth]{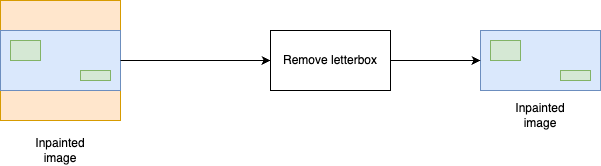}
  \hfill
  \includegraphics[width=0.45\linewidth]{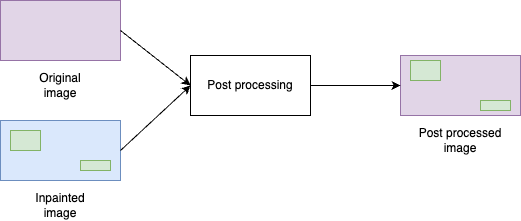}
  \caption{
    Postprocessing workflow: (1.)  
    Removing Letterbox. (2.) Selective Recombination.
  }
  \label{fig:postprocessing_workflow}
  \vspace{-5mm}
\end{figure*}

\vspace{-1mm}
\subsection{In-painting}
\label{sec:inpainting}

To prepare ad images for multilingual localization, the original scene text must be removed without introducing visual artifacts. This step is critical for preserving the overall aesthetic of the advertisement and ensuring that newly translated text can be reimposed on a clean background.

We adopt a Stable Diffusion-based inpainting pipeline tailored for high-resolution ad content. For each image, translatable text regions—previously identified via OCR or human annotation—are used to generate binary masks. These masks indicate areas to be replaced, while unmasked regions are to remain untouched.

Since most diffusion models require fixed-size square inputs (e.g., 512×512 pixels), we first apply letterboxing: padding the image along its shorter dimension with a neutral background to preserve aspect ratio. The padded image and mask are then passed to the inpainting model, which synthesizes visually plausible content conditioned on the surrounding context.

However, direct output from Stable Diffusion may result in hallucinated artifacts outside the intended regions. To mitigate this, we implement a selective recombination step: only the masked (inpainted) patches are extracted and composited back into the original image. This preserves local structure and ensures that background regions not involved in the edit remain visually identical to the original.

Finally, the added padding is removed, restoring the image to its original dimensions. Figures~\ref{fig:inpainting_workflow} and ~\ref{fig:postprocessing_workflow} summarizes the full workflow, including mask generation, letterboxing, inpainting, padding removal, and patch recombination. This pipeline produces high-fidelity backgrounds suitable for typographic reimposition across diverse layouts.

\subsection{Translation of Scene Text}
\label{sec:translation-details}
To perform linguistic conversion of detected textual elements, we use Azure AI Translator to convert translatable text elements into target languages. The service supports over 100 languages and offers both real-time and batch translation through a REST API, making it well-suited for scalable ad localization. Built on transformer architectures and continually updated training data, it provides accurate and context-aware translations compatible with downstream rendering. 

Each translated string is associated with its original bounding box and font metadata for later reimposition responsible for text rendering and layout preservation.

\vspace{-1mm}
\subsection{Localized Text Reimposition}

After translation and inpainting, the final step is to render the translated text back into the image in a way that preserves the original ad’s visual identity. This requires accurately recovering typographic attributes (font family, size, color, and style) and geometric layout properties (position, rotation, and alignment).

To predict the font family, we use a deep classifier based on the EfficientNet-B3~\cite{tan2019efficientnet} architecture. The model is fine-tuned on a synthetically generated dataset comprising over 3,000 font families collected from Google Fonts~\cite{googlefonts}, with samples rendered in diverse resolutions, character combinations, and image augmentations. Input patches from OCR-detected bounding boxes are resized to a fixed resolution and normalized with ImageNet statistics before inference. The classifier outputs a probability distribution over all known font families, and the one with the highest confidence is selected.

\vspace{-1mm}

\begin{algorithm}[t]
\caption{Font Size Estimation and Adaptation}
\label{alg:font-size}

\KwIn{Bounding box diagonal endpoints $(x_1, y_1)$ and $(x_2, y_2)$; original text $t_{\text{orig}}$; translated text $t_{\text{trans}}$}
\KwOut{Adjusted font size $s_t$}

\textbf{Initial Font Size Estimation:}

Compute horizontal length: $w = |x_2 - x_1|$ \;

Compute vertical length: $h = |y_2 - y_1|$ \;

Estimate original font size: $s_o = \min(w, h)$ \;

Set initial font size: $s_t \gets s_o$ \;

Set minimum size threshold: $s_{\min} \gets 0.3 \cdot s_o$ \;

\vspace{-1mm}
\textbf{Dynamic Font Sizing:}

Compute character counts: $n_o = |t_{\text{orig}}|$, $n_t = |t_{\text{trans}}|$ \;

Set step size: $\delta \gets 0.5$ \;

Estimate capacity: $C = \frac{w \cdot h}{s_t^2}$ \;

\If{$n_t > n_o$}{
  \While{$C < n_t$ \textbf{and} $s_t > s_{\min}$}{
    $s_t \gets s_t - \delta$ \;
    $C \gets \frac{w \cdot h}{s_t^2}$ \;
  }
}
\Return{$s_t$} \;
\end{algorithm}
\vspace{-4mm}

Font size is estimated from the bounding box dimensions using the minimum of the horizontal and vertical Euclidean lengths between corners. Since translated text can be longer or shorter than the source, we employ an adaptive sizing strategy: the system calculates the character capacity of the box and iteratively reduces the font size until the translated text fits comfortably, while enforcing a minimum threshold to preserve legibility (see Algorithm 1). This allows flexible resizing without compromising layout fidelity.

\vspace{0.5em}
\noindent
\begin{tcolorbox}[
  floatplacement=!h,
  colback=gray!5!white, 
  colframe=black!75!white, 
  boxrule=0.3mm, 
  arc=1mm, 
  width=\linewidth, 
  fontupper=\ttfamily\small, 
  title=LLM Text Color Output]
\noindent
"name": "Dark Blue",\\
"hex": "\#1E3A8A",\\
"rgb": [30, 58, 138]
\end{tcolorbox}

For font color and style, we leverage large language models (LLMs) such as Claude 3.5~\cite{anthropic2024claude35} and GPT-4o. The annotated ad image—overlaid with translatable bounding boxes—is passed as input to the LLM along with prompts describing context (e.g., product type, brand tone). The model returns structured outputs specifying the color in both RGB and hex formats and style tags such as bold, italic, or underline. We found Claude 3.5 Sonnet provided the most consistent results in capturing brand-consistent typographic cues.

During rendering, we initialize a transparent canvas, render the translated text using the predicted font attributes, and apply rotation if needed. For rotated text, we compute the angle $\theta$ using:
\vspace{-5mm}
\[
\theta = \tan^{-1}\left(\frac{y_2 - y_1}{x_2 - x_1}\right)
\]
where $(x_1, y_1)$ and $(x_2, y_2)$ are the top corners of the bounding box. The rendered text and its mask are rotated accordingly, and we use centroid alignment between the original and rotated regions to ensure spatial accuracy. Horizontal text bypasses the rotation logic and is directly aligned to the top-left anchor.

By integrating deep visual classifiers, adaptive geometry, and LLM-guided styling, our system reimposes translated text with high fidelity to the original appearance—preserving brand identity, layout coherence, and legibility across languages and designs.

%% file: sec/3.5_resultY.tex
\section{Result and Discussion}
The dataset presented in Table 1 was evaluated to measure text recognition accuracy. Levenshtein Distance~\cite{levenshtein1966binary}, Word Error Rate (WER)~\cite{Hunt1990}, Character Error Rate (CER), and F1-score were employed to capture both character- and word-level discrepancies. Averaged across multiple locales, the model achieved a Levenshtein Distance of 0.66, WER of 0.27, CER of 0.05, and an F1-score of 0.90. The low CER indicates strong character-level recognition across languages, while the moderate WER suggests occasional word segmentation issues for non-english locales. The high F1-score reflects balanced precision and recall, underscoring the model’s reliability in multilingual text recognition tasks.

To assess the visual quality of our localized outputs, we conducted a quantitative evaluation using the Learned Perceptual Image Patch Similarity (LPIPS) metric~\cite{zhang2018unreasonable}. LPIPS measures deep feature distance between image pairs and is widely regarded as a reliable proxy for human perceptual similarity.

\begin{figure}[htbp]
 \vspace{-4mm}
  \centering
  \includegraphics[width=0.9\columnwidth]{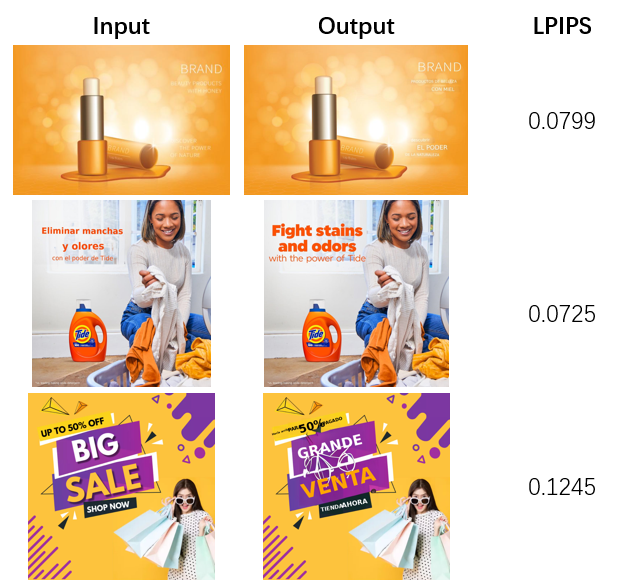}
  \caption{Example of tested Ad samples}
  \label{fig:lpips}
  \vspace{-4mm}
\end{figure}

We selected eight Ad samples across 6 locales (see Figure ~\ref{fig:lpips} for three representative examples) processed by our full pipeline and compared the final localized images to the original, unmasked versions. The LPIPS scores ranged from 0.049 to 0.125, with an average of 0.067, indicating that the inpainting and reimposition process introduced minimal perceptual distortion. While this initial evaluation demonstrates promising perceptual quality, further quantitative benchmarking is planned. We intend to incorporate additional metrics such as IOU and human preference scores to gain a more comprehensive understanding of system performance.

%% file: sec/4_conclusionY.tex
\section{Conclusion}
We present a modular framework for multilingual advertisement localization that integrates scene text detection, background inpainting, machine translation, and layout-aware reimposition. Our system reduces manual effort through automation powered by deep learning and large language models, while preserving visual fidelity and brand identity. Future work includes formal evaluation across six locales, improving support for curved and stylized text, and integrating vision-language models for semantic alignment.